\documentclass[10pt, conference, compsocconf]{IEEEtran}
\ifCLASSINFOpdf
   \usepackage[pdftex]{graphicx}
   \graphicspath{{./figures/}}
\else
\fi
\usepackage[cmex10]{amsmath}
\usepackage{amssymb}
\usepackage{url}
\usepackage{xcolor}

\makeatletter
\newcommand{\thickhline}{%
	\noalign {\ifnum 0=`}\fi \hrule height 1pt
	\futurelet \reserved@a \@xhline
}
\makeatother

\newcommand{\thickcline}[1]{%
	\@thickcline #1\@nil%
}
\usepackage{booktabs}
\begin{document}
\title{Bootstrapping Weakly Supervised Segmentation-free Word Spotting through \\ HMM-based Alignment}

\author{\IEEEauthorblockN{Tomas Wilkinson, Carl Nettelblad}
\IEEEauthorblockA{Department of Information Technology\\
Uppsala University\\
Uppsala, Sweden\\
Email: \{tomas.wilkinson, carl.nettelblad\}@it.uu.se}}

%\author{\IEEEauthorblockN{XXX, YYY}
%\IEEEauthorblockA{XXX\\
%XXX\\
%XXX\\
%XXX}}

\maketitle

\begin{abstract}
Recent work in word spotting in handwritten documents has yielded impressive results. This progress has largely been made by supervised learning systems, which are dependent on manually annotated data, making deployment to new collections a significant effort. In this paper, we propose an approach that utilises transcripts without bounding box annotations to train segmentation-free query-by-string word spotting models, given a partially trained model. This is done through a training-free alignment procedure based on hidden Markov models. This procedure creates a tentative mapping between word region proposals and the transcriptions to automatically create additional weakly annotated training data, without choosing any single alignment possibility as the correct one. When only using between 1\% and 7\% of the fully annotated training sets for partial convergence, we automatically annotate the remaining training data and successfully train using it. On all our datasets, our final trained model then comes within a few mAP\% of the performance from a model trained with the full training set used as ground truth. We believe that this will be a significant advance towards a more general use of word spotting, since digital transcription data will already exist for parts of many collections of interest.
\end{abstract}

\begin{IEEEkeywords}
weakly supervised; segmentation-free word spotting; convolutional neural network; hidden Markov model;
\end{IEEEkeywords}

% \fi
\IEEEpeerreviewmaketitle

\section{Introduction}
Word spotting has enjoyed good progress in the last few years, both in segmentation-based and segmentation-free settings \cite{sudholt2017attribute, wilkinson2018neural, axler2018toward, krishnan2018word}. Much headway is due to the adoption of deep convolutional neural networks by the community. They are powerful models that have come to dominate computer vision over the last few years. The models often require large amounts of annotated data for training. Producing such data can prove expensive and labour-intensive. Tricks like model pre-training and data augmentation have proven useful in reducing the needed data volumes, yet the problem persists. Segmentation-free word spotting in historical manuscripts is a field where annotations can be particularly difficult to get, as expert knowledge is often needed to annotate these manuscripts properly.

For many manuscript collections, there exists an untapped source of partial annotations, in the various transcriptions that have been made throughout the years by scholars as well as enthusiasts directly for research or for the printed publication of older texts \cite{pettersson2015improving}. They are partial in the sense that they only consist of transcriptions, whereas word spotting models typically also require bounding boxes for training. If it would be possible to leverage these transcripts for training word spotting models, that would greatly lessen the labour in switching to a new manuscript collection.

To this end, we propose a novel approach for training segmentation-free query-by-string (SFQBS) word spotting models using only transcripts as supervision. This weakly supervised approach is based on a training-free alignment procedure that uses a hidden Markov model (HMM) to automatically match word region proposals with a manually annotated transcript. This allows us to further improve our models by making use of transcribed data that lack bounding boxes in training. The approach requires a partially trained segmentation-free query-by-string word spotting model, either through pre-training on a different dataset or by fully annotating a few pages. In practice, we manage to use 1-10\% of the full training set for the various datasets. Unlike many alignment methods, the proposed approach does not rely on an explicit line segmentation.

\begin{figure*}[t!]
	\begin{center}
		\includegraphics[width=0.99\linewidth]{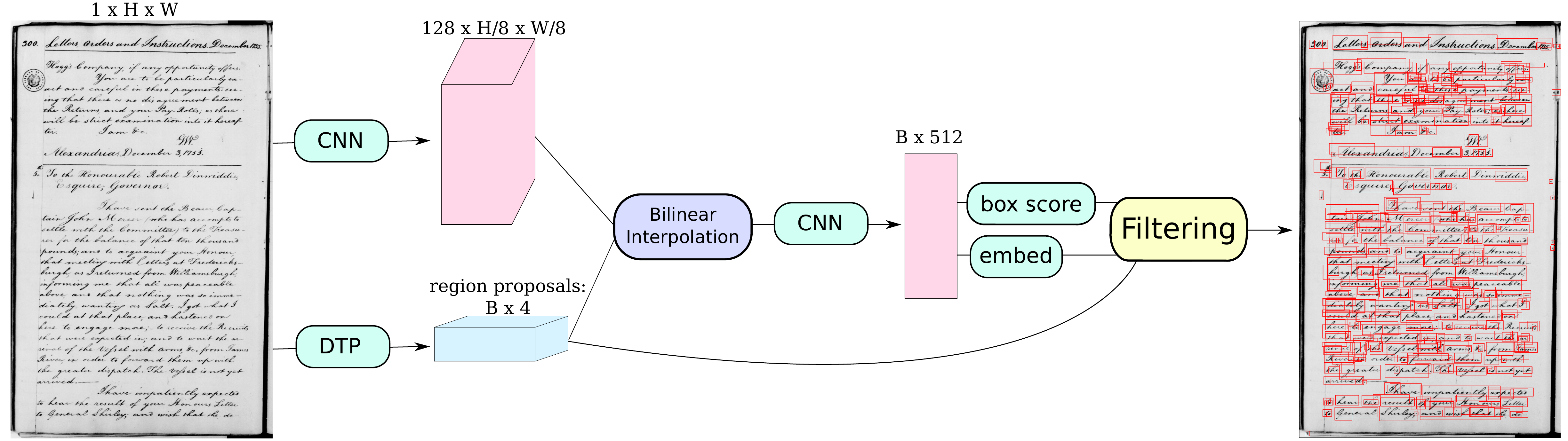}
	\end{center}
	\caption{The Ctrl-F-Mini model: An image is initially fed through the first part of the backbone CNN and region proposals are extracted using the DTP, followed by a bilinear interpolation layer, yielding fixed sized outputs. These are propagated through the rest of the network. Two separate branches compute the scoring and embedding for each region proposal. The score is used i a final proposal filtering step, a two-step process where proposals below a certain threshold are discarded, and the remaining ones are filtered by non-max suppression.}
	\label{fig:model}
\end{figure*}

\section{Related Work}
Automated alignment of transcripts to word images in handwritten manuscripts has a history of over twenty years \cite{tomai2002transcript, kornfield2004text, rothfeder2006aligning, toselli2011alignment, fischer2011transcription, hassner2013ocr}. A popular approach is employing a word or line recognition system to get a transcription and aligning it with the true transcript \cite{tomai2002transcript, rothfeder2006aligning, toselli2011alignment, fischer2011transcription}. Constructing a lexicon for each line using the true transcript, the recogniser \cite{tomai2002transcript} returns a ranked list of possible labels for each word in a line along with a confidence score. Multiple line hypotheses are processed and the most likely line is computed using Dynamic Programming and used for the final mapping. In \cite{fischer2011transcription}, a page-level HMM is created and matched against the sequences of features extracted from the page's text lines. Taking into account spelling variants and word deletions, the HMM-recogniser returns both the optimal word segmentation and word labels. In a similar fashion, the alignment returned by the Viterbi algorithm is used in \cite{toselli2011alignment}.

Another class of alignment methods avoid word recognition. They are based on viewing words in the transcription and word images as sequences and then applying algorithms for sequence alignment \cite{kornfield2004text, hassner2013ocr}. The method detailed in \cite{kornfield2004text} uses Dynamic Time Warping to align word images to the true transcript, relying on a way to measure the dissimilarity between word images and text strings. The features include aspect ratio; width; ascender and descender counts, and are computed on rendered images for the text strings. In \cite{hassner2013ocr}, synthetically generated images were used to match with word hypotheses via dense flow based on local descriptors, avoiding recognition. 

One of the uses of alignment is to annotate data in a manner usable in methods for downstream tasks like handwriting recognition and word spotting, e.g. as in \cite{fischer2011transcription}. Using alignment to directly facilitate learning based on transcription data is closely related to the notion of \emph{weak supervision} \cite{zhou2017brief}, and \emph{inexact supervision} in particular. This is an important line of research in the age of data hungry learning models. 

There has been a prior approach investigating weak supervision in conjunction with word spotting \cite{gurjar2018learning}. However, they define weak supervision as ``training with minimal manual annotation effort'', which is a broader definition than the one from \cite{zhou2017brief}. Their approach is based on pre-training a model on the synthetically generated IIIT-HWS-10k dataset \cite{krishnan2016matching} and then fine tuning with small amounts of real data and seeing how little data they can get away with while maximising performance. This transfer learning based approach differs greatly from the method proposed in this paper as all their training relies on having full annotations, albeit synthetically generated. 

\section{System}
The overall system has two main parts: A segmentation-free query-by-string word spotting model based on deep learning; and an HMM-based alignment module. 

\subsection{Word Spotting Model}
We adopt the Ctrl-F-Mini  model \cite{wilkinson2018neural} using the Discrete Cosine Transform of Words (DCToW) embedding \cite{wilkinson2016semantic} as our word spotting model\footnote{Any word spotting model with the same kind of output is a possible alternative.}. It is a powerful and relatively light-weight model that achieves state-of-the-art results in segmentation-free word spotting. The Ctrl-F-Mini allows for searches using arbitrary text queries in full-page text images, and consists of a pre-activation ResNet34 backbone \cite{he2016identity}; a bilinear interpolation layer that resizes region proposals into a canonical output size \cite{densecap, jaderberg2015spatial}; an embedding network that embeds region proposals into a word embedding space; and external region proposals generated by the Dilated Text Proposals algorithm  \cite{wilkinson2015novel}. The output consists of a set of word proposals and corresponding scores that encode the probability of a proposal being centred on a word. The scores are used for filtering at a later step. See Figure \ref{fig:model} for an overview of the model at test time.

The model takes as input a full page image $I \in \mathbb{R}^{1 \times H \times W}$, and region proposals $R \in \mathbb{R}^{B \times 4}$. The image $I$ is fed through half the ResNet, until it has been downsampled by a factor $8$ in each dimension, resulting in feature maps $M \in \mathbb{R}^{C \times H/8 \times W/8}$. Then the overlap between the region proposals and ground truth boxes is computed, and $128$ proposals whose overlap is greater than $75\%$ are randomly chosen as positive training examples $P \in \mathbb{R}^{128 \times 4}$. Those under $40\%$ overlap are defined as negative examples $N \in \mathbb{R}^{128 \times 4}$ from which another $128$ are sampled, and those in between are discarded. Then $M, P,$ and $N$ are fed into the bilinear interpolation module. Here, the differently sized proposals are all mapped to a fixed size output. They are then fed through the rest of the ResNet and into two parallel branches. The first is a scoring branch that tries to estimate the probability of a proposal being centred on a word. The second is a small fully connected network with two hidden layers, each with $4096$ neurons, and an output dimensionality of $108$, matching that of the DCToW.

Two loss functions are used, a cross entropy loss $L_{score}$ for the scoring branch, and a cosine loss for the embedding branch. For prediction $x$ and ground truth embeddings $y$, the cosine loss is defined as
\begin{equation}\label{eq:cosine_loss}
L_{emb}(x, y) = \frac{1}{n}\sum\limits_{i=1}^n (1 - \frac{x_i^\mathsf{T}y_i }{||x_i|| \cdot ||y_i||})
\end{equation}
where $n$ is the batch size. The total loss is 
\begin{equation}
	L_{total} = 0.1\cdot L_{score} + 3 \cdot L_{emb}
\end{equation}

\subsubsection{Searching}
For searching in a new manuscript collection, the first step is to extract region proposals. Then, we can run the model as described in Figure \ref{fig:model}, extracting scores and embeddings for the proposals, and filtering them by score thresholding and non-max suppression. Score thresholding entails discarding proposals with a score lower than a particular threshold. Non-max suppression is a greedy algorithm that, for a set of proposals with overlap greater than a threshold, selects the proposal with the highest score and removes the rest. The remaining proposals and embeddings constitute the database within which searches are performed.

Once region proposals, scores and embeddings for a page have been extracted, it is possible to search for any text string. Given a query, the search procedure starts with computing the cosine similarity between the query and the database and doing another non-max suppression filtering with the cosine similarity as a score, this time allowing zero overlap between overlapping proposals. Then filtered results are sorted according to the similarity to the query.

\subsection{Alignment Module}
For the alignment module, we chose to use an approach based on hidden Markov models (HMMs). These have been used for various alignment tasks for decades \cite{rabiner1989tutorial}. The model consists of two sequences: the observed symbols and the hidden sequence of states. Between a state $i$ and a state $j$ there is a transition probability $\varphi_{ij}$. For each state $i$ and symbol $k$ there is an emission probability $\theta_{ik}$. Depending on the desired usage mode, one can identify the single most likely state sequence based on a set of observations (the Viterbi algorithm), identify the marginalized state probability distribution for each observation over all possible sequences (the forward-backward algorithm), and also allow fitting of one or more parameters using observed sequences.

In our setting, the symbols in the observed sequence consisted of the unique words on the page. The hidden states consisted of a subset of candidate boxes identified using the word-spotting model. For each box $i$ and ground-truth word $k$ with prediction and ground-truth $x, y$ as in eq. \ref{eq:cosine_loss}, the emission likelihood is defined based on the supposed null distribution for the cosine similarity. With a 108-dimensional space, the variance for the scalar product of two independent random unit vectors is $1/108$, and the dimensionality makes the distribution close to normal. Thus, we use a scaled Gaussian to represent our likelihoods:
\begin{equation} \label{eq:emission}
\theta_{ij} = \exp \left ( \left ( \frac{x_i^\mathsf{T}y_k }{||x_i|| \cdot ||y_k||} - 1 \right )^2 \frac{{ 108 }}{2}\right )
\end{equation}

We use likelihoods for clarity to avoid unnecessary renormalization (emission probabilities implicitly always sum to 1). It can also be noted that for ground-truth words that occur multiple times on a page, the emission probabilities will be identical. The emission probabilities only propagate the insight from the word-spotting model of the similarity between box contents and the word strings.

The transition probabilities impose the main rules that successive words on the same line in a left-to-right, top-to-bottom script occur further to the right at approximately the same height, and that words on a following line occur below the previous line. Furthermore our likelihood includes penalties for overlapping boxes, and boxes with gaps between them when on the same line. These two penalties ensure that alignments based on non-sensical segmentations are avoided if there is another reasonable alignment.

Specifically, if we denote the states $i$ and $j$ we can define the rule part of the likelihoods at the current position $k$ in the sequence as
\begin{equation} \label{eq:transitionrule}
	\varphi^r_{ij} = \begin{cases}
		1 & \lnot N \land r_j > l_i \land t_i - h < t_j \land t_i + h < t_j \\
		1 & N \land b_j > t_i \\
		\varepsilon & \mbox{otherwise}
	\end{cases}
\end{equation}
where the box extremes are $l,r,t,b$ (left right top bottom), $h = \max(b_i - t_i, b_j - t_j)$ is the line height, $N$ indicates whether there is a line break between the two words in the transcription at the current position in the sequence, and $\varepsilon$ is a small value for residual cases ($0.01$ is a reasonable choice in practice). The penalties are defined as:
\begin{equation} \label{eq:transitionpen}
\begin{aligned}
\varphi^p_{ij} = &\left (1 - \frac{(1 - \varepsilon) A_{\cap i, j}}{\min(A_i, A_j)} \right ) \cdot \\
& \begin{cases}
\varepsilon + (1 - \varepsilon)\frac{A_i + A_j - A_{\cap i, j}}{A_{\cup i, j}} & \lnot N\\
1 & N\\
\end{cases} 
\end{aligned}
\end{equation}
where $A_i$ is the area of box $i$, $\cap$ indicates the intersection and $\cup$ indicates the bounding rectangle including both boxes. The total transition likelihood is then $\varphi_{ij} = \varphi^r_{ij} \varphi^p_{ij}$.

Using this model and the forward-backword algorithm, posterior likelihoods $\pi_{ik}$ can be computed for all states $i$ and positions $k$. Based on the properties of the hidden Markov model, these likelihoods will take into account any possible alignment and retain memory over the full sequence. Box assignments that do not result in a valid overall alignment will be penalised by the $\varepsilon$ factor one or multiple times. The fact that the forward-backward algorithm evaluates all possible alignments, not only the single most likely one, allows for reasonable results even if grave mistakes are present in parts of the text.

%By normalizing $\pi_{ik}$ and $\theta_{ik}$, a desired change to the original cosine embedding scores can be produced by computing $\log{\frac{pi_{ij}}{\theta_{ik}}}$. This information can then be fed back to the word spotting model as a gradient.

\subsubsection{Alignment Procedure}
Given a transcription, we use the word-spotting model to search for each word in the transcription in reading order, and save the result in the form of a sorted list of boxes and scores. The alignment module matches the transcriptions with the boxes and returns resulting probabilities after alignment. With this, we add the weakly supervised data to our original data and further train our word spotting model. As for computational effort, most of it is spent on searching for the words in the transcription, as the HMM model evaluation itself is extremely fast. 
%For flexibility in the implementation, this is currently done in the form of separate text files. The HMM model evaluation itself is extremely fast. The computational time spent on input file parsing is comparable to or actually exceeds the time spent evaluating the model.

\section{Experiments}
We conduct two kinds of experimental evaluations of our approach. The first is a direct evaluation of the alignments produced by our proposed approach. Note that we do not compare to other alignment methods as this is outside the scope of the paper, we merely want to demonstrate that we get reasonable alignment performance and that this level of performance proves adequate to inform our weakly supervised training. The second experiment evaluates the performance of the word spotting model when using our proposed weakly supervised training approach.

\subsection{Datasets}
We evaluate our approach across three handwriting datasets, two created from historical manuscript collections and one modern. 

The \textbf{George Washington} (GW) dataset \cite{lavrenko2004holistic} contains $20$ pages from Series 2, Letterbooks of the George Washington papers\footnote{\url{https://www.loc.gov/collections/george-washington-papers/about-this-collection/}}. They were written in English in the mid 18\textsuperscript{th} century by George Washington and his secretaries. The text of the pages in the dataset is homogeneous enough to often be considered a single writer dataset. We follow the evaluation procedure detailed in \cite{rothacker2015segmentation}, where the 20 pages are split into a $15$ page training and validation split, and $5$ pages for testing. Evaluations are performed using $4$-fold cross validation, and reported results are their average. For the weak evaluation, we use $1$ page ($7\%$) of the regular training split for the initial training and only the transcriptions on the remaining $13$ pages.

The \textbf{Barcelona Historical Handwritten Marriages } (BH2M) dataset \cite{BH2M2014} is composed of 174 pages from a book of marriage records written in old Catalan between $1617$ and $1619$. The text is 
written by a single hand, but presents problems such as crossed over words and paragraphs; drawings interspersed among the text; and words written between text lines. The recommended split proposed in \cite{BH2M2014} is $100$ pages for training, $34$ for validation and $40$ for testing. We use $10$ pages ($10\%$) for the initial training.

The \textbf{Offline IAM} dataset \cite{marti2002iam} is a modern, cursive, multi-writer dataset written in English. It consists of 1539 pages, written by 657 writers. We follow the splits detailed in \emph{Large Writer Independent Text Line Recognition Task}, which has no writer overlap between the different splits. In line with \cite{almazan2014word}, we remove queries that come from segmentations marked as erroneous. We also remove stop words from the set of queries used for evaluation. Due to model constraints, we further remove ground truth boxes that are so small that they collapse to width or height of zero when downsampled by a factor 8. As with the BH2M dataset, we use $10$ pages (around $1\%$) for the initial training.

%\footnote{\url{https://github.com/almazan/watts/blob/master/data/swIAM.txt}} 

\subsection{Evaluation} \label{sec:evaluation}
To evaluate the word spotting we adopt the standard metric Mean Average Precision (mAP). For a database of size $N$, the Average Precision is defined as
\begin{equation}
	\textnormal{AP}(q) = \frac{1}{R_q} \sum\limits_{k=1}^N P_k \cdot r_k
\end{equation}
where $R_q$ is the number of relevant results for query $q$, $P_k$ is the precision measured up to rank $k$ in the results, and $r_k$ is the relevance of the $k^{\textnormal{th}}$ retrieved result. The $k^{\textnormal{th}}$ result is considered relevant ($r_k = 1$) if it matches the query and the IoU overlap is greater than a threshold $t_o \in \{0.25, 0.5\}$, otherwise $r_k = 0$. The mAP is then calculated as the average over all queries 
\begin{equation}
\textnormal{mAP} = \frac{1}{Q} \sum\limits_{q=1}^Q AP(q) 
\end{equation}
where $Q$ is the number of queries. All the unique ground truth labels from the test set are used as queries.

To evaluate the alignment we define two sets, $P^o$ and $P^t$. The first contains the $n$ aligned bounding boxes $B \in \mathbb{R}^{n \times 4}$ that overlap with the $m$ ground truth boxes greater than $50\%$.  $P^t$ contains the aligned boxes whose labels match the label of the ground truth bounding box that it maximally overlaps with. Putting these together, we get the transcriptions accuracy as
\begin{equation}
\textnormal{acc}(B) = \frac{1}{\max(m, n)} \sum\limits_{i=1}^n \mathbb{I}(b_i \in P^o \land b_i \in P^t)
\end{equation}
where $\mathbb{I}$ is the indicator function that evaluates to 1 when its arguments are true and zero otherwise. Note that this metric does not penalise matching multiple proposals to the same ground truth box. For training word spotting models, this is no issue as we only need proposals to have correct label and sufficient overlap. Multiple proposals being assigned the same label would only yield more diverse training data.

\subsection{Training}\label{sec:training}
First, we train a model (Ctrl-F-Mini) on the IIIT-HWS-10k dataset \cite{krishnan2016matching}. This is used to initialise the initial model for each dataset. Then, a model is trained on the reduced training split for 10k iterations (learning rate of $10^{-3}$). After that, the weak supervision commences by initialising with the seed model and performing alignment on the remaining training dataset. This weakly supervised model is also trained for 10k iterations (learning rate of $10^{-4}$). For multiple rounds of alignment, the weakly trained model from the previous round is used for the alignment and trained for another 10k iterations each realignment. Weights are updated using ADAM \cite{kingma2014adam}.

During training, models are evaluated on the validation set every $1000$ iterations and the best model is kept. Furthermore, all models are trained using on-the-fly augmentation based on the in-place scheme from \cite{wilkinson2018neural}. Full-page augmentation was avoided due to the computational overhead of re-extracting region proposals on the fly.
%, whereas in-place augmentation can reuse proposals extracted for the original images.

\subsection{Results}
The first set of experiments evaluate the alignment performance for using the metrics described in Section \ref{sec:evaluation}, see Table \ref{tab:alignment_results}. The baseline model is trained only on the fully annotated data. The supervised model is trained using the ground truth alignment, i.e. the regular supervised learning setting. This acts as an upper baseline for our alignment approach. The aligned model is trained using the weak supervision, and for aligned x2 and x3, $1$ and $2$ extra rounds of alignment are done, as detailed in Section \ref{sec:training}. Both the baseline and supervised models are trained with the same amount of iterations as the aligned x3 model. 

We first note that the alignment performance is over $90\%$ for all entries. For the Washington and BH2M datasets the alignment performance drops slightly, possibly due to drift of the models trained on the weakly annotated data. There is a more noticeable difference for IAM, halving the error from the baseline to the aligned model. The accuracy is further increased for further rounds of alignment. In addition, for IAM we come close to achieving the same accuracy as the fully supervised model.
\begin{table}[t!]
	\renewcommand{\arraystretch}{1.3}
	\caption{Alignment accuracy (in \%) for the different models.}
	\label{tab:alignment_results}
	\centering
	\begin{tabular}{lccc}
		& Washington & BH2M & IAM \\ 
		Model &  &  & \\
		\thickhline
		baseline & 91.4 & 93.7 & 91.2 \\
		supervised & 96.9 & 93.6 & 98.0 \\
		aligned & 90.7 & 91.8 & 95.7 \\
		aligned x2 & 90.7 & 91.3 & 96.3 \\
		aligned x3 & 91.0 & 90.6 & 96.6 \\
		\thickhline
	\end{tabular}
\end{table}

In a second set of experiments, we measured the performance of the different models in terms of mAP (shown in Table \ref{tab:map_results}). We can make a few observations from the experiments. First, all aligned models perform better than their respective baselines. Second, further rounds of alignment generally give improved results, though with diminishing returns. Third, models are quite close to achieving the same mAP as the fully supervised model, i.e., the upper baseline. For reference we also include IIIT-HWS-10K pre-trained model and the best published results.

\section{Discussion}
The results of Table \ref{tab:alignment_results} show that it is possible to use only small amounts of data in order to get a good enough model for a highly accurate alignment. This is especially true for IAM, where only $1\%$ of the regular training split was used for the initial training. This indicates that it might be possible to fully circumvent the use of fully annotated data, perhaps by relying on synthetically generated data like the IIIT-HWS-10k dataset, or transfer learning from another dataset. Furthermore, since we use a forward-backward approach, rendering box probabilities, rather than the Viterbi approach of a singel correct alignment, even if the single most likely alignment is incorrect, the correct box for a certain word is frequently given a substantial likelihood boost.

In Table \ref{tab:map_results}, the weakly trained models outperform the lower baseline model across all three datasets. Furthermore, on the span between the lower and upper baselines (supervised model), the aligned models perform closer to the upper end. This result indicates that our procedure for weakly supervised training can be readily applied to many settings with promising results, which could most likely be improved through further refinement. Unfortunately, to our knowledge there are no other published methods for weakly supervised word spotting that are comparable for us to evaluate our approach against.

\begin{table}[h]
	\renewcommand{\arraystretch}{1.3}
	\caption{Query-by-String mAP (in \%) for the different models evaluated with $50\%$ and $25\%$ overlap thresholds.}
	\label{tab:map_results}
	\centering
	\begin{tabular}{lcccccc}
		
		& \multicolumn{2}{c}{Washington} & \multicolumn{2}{c}{BH2M} & \multicolumn{2}{c}{IAM} \\ \cmidrule(r){2-3}\cmidrule(lr){4-5}\cmidrule(lr){6-7}
		Model & 50\% & 25\% & 50\% & 25\% & 50\% & 25\% \\
		\thickhline	
		IIIT-HWS-10K pretrained & 14.4 & 20.7 & 15.9 & 20.6 & 17.3 & 18.5 \\
		\thickhline
		baseline & 79.8 & 83.1 & 79.1 & 81.7 & 51.9 & 52.7\\
		supervised & 90.7 & 93.9 & 83.5 & 87.5 & 81.6 & 82.7\\
		aligned & 88.1 & 92.2 & 80.9 & 85.1 & 71.1 & 72.3\\
		aligned x2 & 87.9 & 92.4 & 81.6 & 85.8 & 73.3 & 74.6\\
		aligned x3 & 88.1 & 92.0 & 81.4 & 86.2 & 75.6 & 77.0\\
		
		\thickhline
		Ctrl-F-Mini \cite{wilkinson2018neural} & 93.5 & 96.3 & - & - & 84.9 & 86.8 \\
		Encoder-Decoder Net \cite{axler2018toward} & - & - & - & - & 85.4 & 85.6\\
		\thickhline
	\end{tabular}
\end{table}

Multiple rounds of alignment in a very naive feedback loop showed improved results, or maintained, results for all datasets, with the most clear improvements when initial results were low. More advanced bootstrapping or feedback schemes should be possible.

\section{Conclusion}
We introduce a weakly supervised training approach for SFQBS word spotting models whose core part is a training-free HMM-based alignment module. This allows us to train without expensive bounding box annotations, relying only on transcripts. This is important as there exist many manuscript collections where sections or excerpts of various length have been transcribed. These can now be used to train word spotting models. Depending on the dataset, we use as little as $1-10\%$ of the regular training data to bootstrap the weakly supervised learning process. Across all experiments, we outperform the baseline of not including the weak data, and manage to come within a few points of mAP to a upper baseline model trained on the complete, and fully annotated training data. We hope that this work will help in the adoption of word spotting into manuscript-based research by lowering the deployment costs to new manuscript collections.

\section*{Acknowledgment}
This project is a part of q2b, From quill to bytes, which is a digital humanities initiative sponsored by the Swedish Research Council (Dnr 2012-5743), Riksbankens Jubileumsfond (NHS14-2068:1) and Uppsala University.



\bibliographystyle{IEEEtran}
%\begin{thebibliography}{1}
%\end{thebibliography}

\bibliography{weak_training}

\end{document}